\documentclass[11pt,a4paper]{article}
\usepackage[hyperref]{naaclhtl2019}
\usepackage{times}
\usepackage{latexsym}
\usepackage{url}
\usepackage{textcomp}
\usepackage{bbm}
\usepackage{amsmath}
\usepackage{booktabs}
\usepackage{tabularx}
\usepackage{graphicx}
\usepackage{dialogue}
\usepackage{mathtools}
\usepackage{hyperref}
\usepackage{enumitem}
\setitemize{noitemsep,topsep=5pt,parsep=5pt,partopsep=5pt}

\usepackage{multirow}
\usepackage{mdframed}
\usepackage{tcolorbox}

\usepackage{xcolor,pifont}

\mdfdefinestyle{dialogue}{
    backgroundcolor=yellow!20,
    innermargin=5pt
}

\usepackage{amssymb}
\usepackage{soul}
\makeatletter
\def\blfootnote{\xdef\@thefnmark{}\@footnotetext}
\makeatother

\newcommand{\quac}[0]{\daffy}
\newcommand{\quact}[0]{QuAC}
\newcommand{\coqa}[0]{CoQA}
\newcommand{\squad}[0]{SQuAD~2.0}

\newcommand{\daffy}[0]{\includegraphics[width=.04\textwidth]{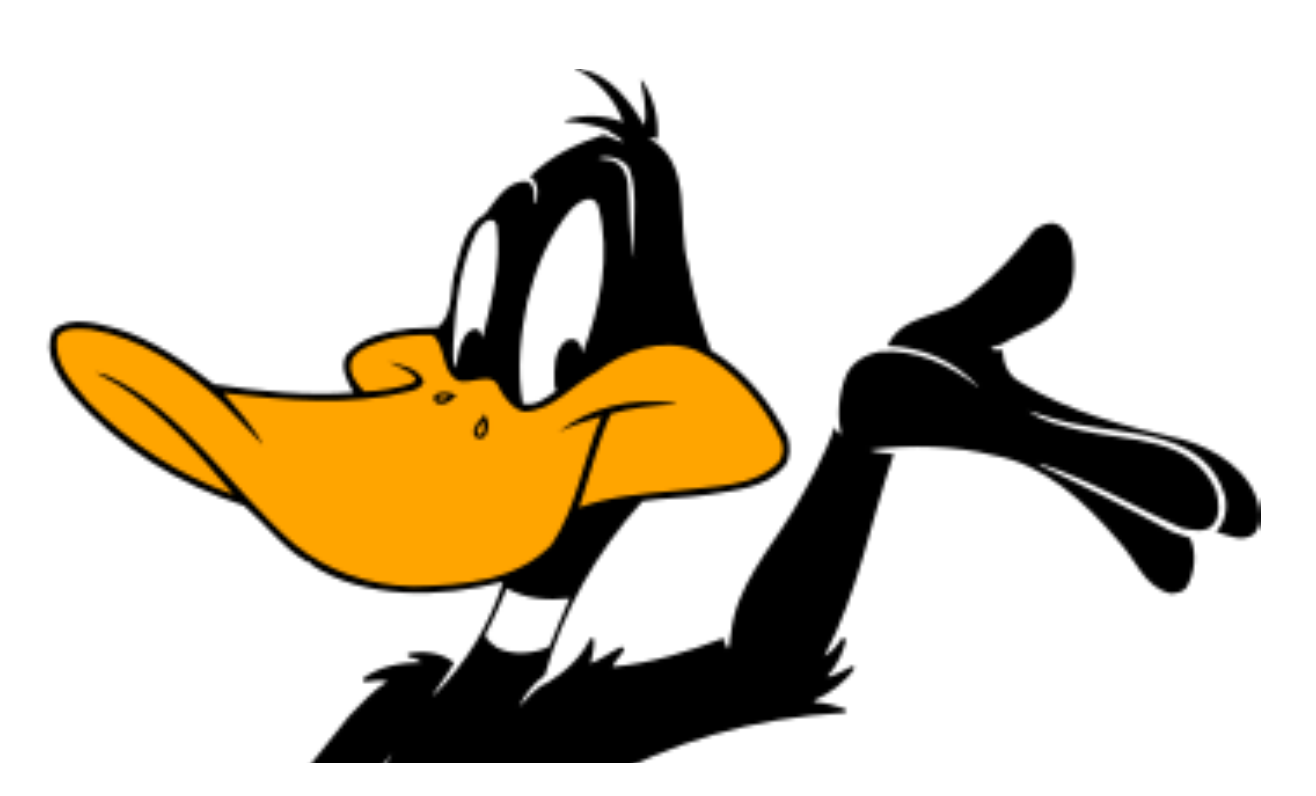}}

\newcolumntype{L}[1]{>{\raggedright\let\newline\\\arraybackslash\hspace{0pt}}m{#1}}
\newcolumntype{C}[1]{>{\centering\let\newline\\\arraybackslash\hspace{0pt}}m{#1}}
\newcolumntype{R}[1]{>{\raggedleft\let\newline\\\arraybackslash\hspace{0pt}}m{#1}}

\title{A Qualitative Comparison of CoQA, SQuAD 2.0 and \quac QuAC}

\author{
\textbf{Mark Yatskar} \vspace{2mm} \\Allen Institute for Artificial Intelligence\\ \texttt{marky@allenai.org}
}
\aclfinalcopy

\begin{document}
\maketitle
\begin{abstract}
We compare three new datasets for question answering: \squad, \quact, and \coqa, along several of their new features: (1) unanswerable questions, (2) multi-turn interactions, and (3) abstractive answers.
We show that the datasets provide complementary coverage of the first two aspects, but weak coverage of the third.
Because of the datasets' structural similarity, a single extractive model can be easily adapted to any of the datasets and we show improved baseline results on both \squad~ and~\coqa.
Despite the similarity, models trained on one dataset are ineffective on another dataset, but we find moderate performance improvement through pretraining.
To encourage cross-evaluation, we release code for conversion between datasets at \href{https://github.com/my89/co-squac}{https://github.com/my89/co-squac}.
\end{abstract}
\begin{table*}[th]
 \small
 \centering
\begin{tabular}
{c|c|c|c|c|c|c|c}

\multirow{2}{*}{Dataset}& \multirow{2}{*}{\parbox{1cm}{\centering Entity Salad}} &\multirow{2}{*}{\parbox{1cm}{\centering False Premise}} & \multirow{2}{*}{\parbox{1.2cm}{\centering Topic Error}} &
\multirow{2}{*}{\parbox{1.8 cm}{\centering Missing Information}}  & \multirow{2}{*}{\parbox{1.7cm}{\centering Content Negation}} &   
\multirow{2}{*}{\parbox{1.7cm}{\centering Answerable Questions}} &   
\multirow{2}{*}{\parbox{1.7cm}{\centering Total Questions}} 
\\
&&&&&\\\midrule
\coqa & 0.0 & 0.0  &  0.0 &  60.0 & 0.0 & 40.0 & 5 (0.5\%)\\
\squad & 21.3 & 21.3 & 13.5 & 16.1 & 16.1 & 10.9 & 230 (50.1\%) \\
\quact & 5.5 & 0.0 & 16.4 & 71.2 & 0.0 & 6.8 & 73 (20.2\%)\\ 
 
\end{tabular}
\vspace{-7pt}
\caption{Comparison of unanswerable questions on 50 random contexts from the development set of each dataset. \squad~contains a diverse set of circumstances that make questions unanswerable, \quact~focuses on information that could plausibly be in context material and \coqa~does not significantly cover unanswerable questions.} 
 \label{tab:unanswerable}
\vspace{-7pt}
\end{table*}

\section{Introduction}
\label{sec:introduction}
Question answering on textual data has served as a challenge problem for the NLP community \cite{voorhees2001trec,mctest}. 
With the development of large scale benchmarks and sufficiently simple evaluations ~\cite{trischler2016newsqa,Nguyen2016MSMA,hermann2015teaching} progress has been rapid. 
In recent evaluation on 
SQuAD ~\cite{rajpurkar2016squad}, performance exceeded that of annotators ~\cite{wang2018multi,hu2017reinforced,wang2017gated}. 

In response to this development, there have been a flurry of new datasets. 
In this work, we analyze three such new proposed datasets, \squad~\cite{Rajpurkar2018KnowWY}, \quac\quact~\cite{ChoiQuAC2018}, and \coqa~\cite{Reddy18}.\footnote{A review of other new datasets is in the related work.}
In each of these datasets, crowd workers are asked to (1) produce questions about a paragraph of text (context) and (2) produce a reply by either indicating there is no answer, or providing an extractive answer from the context by highlighting one contiguous span.
QuAC and CoQA contain two other features: questions are asked in the form of a dialog, where co-reference to previous interactions is possible and directly answering yes/no is possible.
CoQA also allows workers to edit the spans to provide abstractive answers.\footnote{Also, \squad~and \quact~ cover only Wikipedia text, \coqa~ covers six other domains and \quact~is the only one of these datasets that doesn't allow the questioner to see the context before formulating a question.} 

We compare these three datasets along several of their new features: (1) unanswerable questions, (2) multi-turn interactions, and (3) abstractive answers.
Unanswerable question coverage is complementary among datasets; 
\squad~focuses more on questions of extreme confusion, such as false premise questions, while \quact~primarily focuses on missing information.
\quact~and \coqa~ dialogs simulate different types of user behavior: \quact~dialogs often switch topics while \coqa~dialogs include more queries for details.
Unfortunately, no dataset provides significant coverage of abstractive answers beyond yes/no answers, and we show that a method can achieve an extractive answer upper bound of 100 and 97.8 F1 on \quact~ and \coqa~, respectively.

Motivated by the above analysis, we apply the baseline presented in \quact~\cite{ChoiQuAC2018}, BiDAF++, a model based on BiDAF~\cite{seo2016bidirectional}, augmented with self attention~\cite{clark2017simple} and ELMo contextualized embeddings~\cite{peters2018deep} to all datasets.
Experiments show that this extractive baseline outperforms existing extractive and abstractive baselines on \coqa~by 14.2 and 2.7 F1 respectively.
Finally, we show models can transfer between datasets with pretraining yielding moderate gains.
\footnote{To facilitate easy future cross-evaluation, we release tools for conversion between these dataset.}

\begin{table*}[t]
 \small
 \centering
\begin{tabular}
{c|c|c|c|c|c|c|c}

\multirow{2}{*}{Dataset}& \multirow{2}{*}{\parbox{1cm}{\centering Topic Shift}} &\multirow{2}{*}{\parbox{1cm}{\centering Drill Down}} & \multirow{2}{*}{\parbox{1.2cm}{\centering Return to Topic }} &
\multirow{2}{*}{\parbox{1.8 cm}{\centering Clarification Question}}  & \multirow{2}{*}{\parbox{1.7cm}{\centering Definition Question }} &
\multirow{2}{*}{\parbox{1.7cm}{\centering Sentence Coverage }} &
\multirow{2}{*}{\parbox{1.7cm}{\centering Total Questions }} 
\\
&&&&&\\\midrule
\coqa & 21.6 & 72.0  & 2.9 & 0.0 & 0.7  & 63.3 & 722 \\
\quact & 35.4 & 55.3 & 5.6 & 0.7 & 3.0  & 28.4 & 302  \\ 
 
\end{tabular}
\vspace{-5pt}
\caption{Comparison of dialog features in 50 random contexts from the development set of each dataset. \coqa~ contains questions that drill into details about topics and cover 60\% of sentences in the context while in \quact~ dialog switch topic more often and cover less than 30\% of sentences. Neither dataset has a significant number of returns to previous topics, clarifications, or definitional interactions. } 
\vspace{-5pt}
 \label{tab:dialog}
\end{table*}

\begin{table*}[t]
 \small
 \centering
\begin{tabular}
{c|c|c|c|c|c|c}

\multirow{1}{*}{Dataset}& 
\multirow{1}{*}{\parbox{1cm}{\centering Yes/No}} 
& \multirow{1}{*}{\parbox{1cm}{\centering Coref }} 
& \multirow{1}{*}{\parbox{1.8 cm}{\centering Counting}} 
& \multirow{1}{*}{\parbox{1.7cm}{\centering Picking }} &
\multirow{1}{*}{\parbox{1.7cm}{\centering Fluency }} &
\multirow{1}{*}{\parbox{1.7cm}{\centering Max F1 }} 
\\
\midrule
\coqa & 21.4 & 3.2 & 1.3 & 0.6 & 4.2 & 97.8  \\ 
\quact  & 21.1 & 0.0 & 0.0 & 0.0 & 0.0 & 100.0 \\
 
\end{tabular}
     \vspace{-5pt}

\caption{Comparison of abstractive features in 50 random contexts in the develoment set of each dataset. Both \quact~and \coqa~contain yes/no questions while \coqa~also contains answers that improve fluency through abstractive behavior. The extractive upper bound from \coqa~is high because most absractivive answers involve adding a pronoun (Coref) or inserting prepositions and changing word forms (Fluency) to existing extractive answers, resulting in extremely high overlap with possible extractive answers.} 
 \label{tab:abstractive}
     \vspace{-7pt}

\end{table*}

\section{Dataset Analysis}
\label{sec:data_analysis}
In this section we analyze unanswerable questions, dialog features, abstractive answers in \squad,~\quact, and~\coqa. 
All analysis was performed by the authors, on a random sample of 50 contexts (300-700 questions) from the development set of each dataset. 

 \vspace{-5pt}
\subsection{Unanswerable Questions}
\label{sec:unanswerable}
In Table~\ref{tab:unanswerable} we compare types of unanswerable questions across dataset.
We identify five types of questions found between the datasets:
    \vspace{-7pt}
    \paragraph{1. Entity Salad} A nonsensical reference to entities found in the context or made-up entities (e.g. {\it ``What infinite hierarchy implies that the graph isomorphism problem s NQ-complete?''}). Such questions are unanswerable for any context.
    \vspace{-7pt}
    \paragraph{2. False Premise} A fact that contradicts the context is asserted in the question (e.g. {\it ``When is the correlation positive?''} but in the context says {\it ``the correlation is strictly negative''}). 
    \vspace{-7pt}
    \paragraph{3. Topic Error} A questions that references an entity in the context but the context does not focus on that entity (e.g {\it ``How many earthquakes occur in California?''} when the article focus is actually about {\it ``Southern California''} ). Such questions potentially have answers, but it would be unlikely for the answer to be found in the context. 
    \vspace{-7pt}
    \paragraph{4. Missing Information} A question who's answer could be plausibly in the context but is not (e.g. {\it ``What is the record high in January?''} and the article is about temperature extremes). Such questions have an answer but it is not mentioned. 
    \vspace{-7pt}
    \paragraph{5. Content Negation} A question which asks for the opposite information of something mentioned in the context (e.g.  {\it ``Who {\bf didn’t} cause the dissolution of the Holy Roman Empire?''}). Such questions either have answers that are the set of all entities other than the one mentioned or answers that could be found in some other context.

\paragraph{Results}
\squad~contains the highest diversity of unanswerable questions of all datasets analyzed. 
Some \squad~questions are unlikely to be asked without significant foreknowledge of the context material and do not occur in \quact.~\footnote{Such questions resemble text from entailment datasets such as SNLI~\cite{bowman2015large} and seem more likely to arise if questioners are receiving very complex information and become confused.}
Both \squad~ and \quact~ cover a significant number of unanswerable questions that could be plausibly in the article.
The difference in settings and distributions of unanswerable questions in \squad~ and \quact~ appear to be complementary: \squad~ focuses more on questions simulating questioner confusion, while \quact~ primarily focuses on missing information.~\footnote{\coqa~ does not contain a significant number of unanswerable questions, and many of the ones that do exist are erroneously marked.}


\subsection{Dialog Features}
\label{sec:dialog}
In Table~\ref{tab:dialog} we analyze five dialog behaviors:
    \vspace{-7pt}
    \paragraph{1. Topic Shift} A question about something previously discussed (e.g. ``Q: How does he try to take over? ... Q: Where do they live?'').
    \vspace{-7pt}
    \paragraph{2. Drill Down} A request for more information about a topic being discussed (e.g. ``A: The Sherpas call Mount Everest Chomolungma. Q: Is Mt. Everest a holy site for them?'')
    \vspace{-7pt}
    \paragraph{3. Topic Return} Asking about a topic again after it had previously been shifted away from.
    \vspace{-7pt}
    \paragraph{4. Clarification} Reformulating a question that had previously been asked. 
    \vspace{-7pt}
    \paragraph{5. Definition} Asking what is meant by a term (e.g. ``What are polygenes?'')


\paragraph{Results} \quact~and \coqa~ contain many similar features but at very different rates, offering complementary coverage of types of user behavior. 
\coqa~ dialogs drill down for details significantly more frequently and cover more than 60\% of sentences in the context material (Sentence Coverage). 
\quact~dialogs shift to new topics frequently and cover less than 30\% of sentences in the context.
Both datasets contain only a small numbers of definition questions and returns to previous topics and few requests for clarification.
\subsection{Abstractive Answers}
\label{sec:abstraction}
Table~\ref{tab:abstractive} compares abstractive behavior in \coqa~and \quact. We observed five phenomena: 
    

\vspace{-7pt}
\paragraph{1. Yes/No} Questions annotated with yes/no. In \quact~such questions and their corresponding yes or no are marked in addition to an extractive answer. In~\coqa, the single token ``yes'' or ``no'' is simply asserted as the abstractive answer, with an extractive answer provided in the rationale (e.g. ``Q: Is atmosphere one of them? A: yes'').
\vspace{-7pt}
\paragraph{2. Coref} Coreference is added to previously mentioned entities in either context or question (e.g. ``Q: How was France's economy in the late 2000s? A: {\bf it} entered the recession''). 
\vspace{-7pt}
\paragraph{3. Count} Counting how many entities of some type were mentioned (e.g. ``Q: how many specific genetic traits are named? A: five'')
\vspace{-7pt}
\paragraph{4. Picking} A question that requires the answer to pick from a set defined in the question (e.g. ``Q: Is this a boy or a girl? A: boy)
\vspace{-7pt}
\paragraph{5. Fluency} Adding a preposition, changing the form of a word, or merging two non-contiguous spans (e.g. ``Q: how did he get away? A: {\bf by} foot) 

\paragraph{Results} Both \quact~and \coqa~have a similar rate of yes/no questions. \quact~contains no other abstractive phenomena while \coqa~contains a small number of predominately insertions, often at the beginning of an extractive span, for coreference and or other fluency improvements. 
Because abstractive behavior in \coqa~ includes mostly small modifications to spans in the context, the maximum achievable performance by a model that predicts spans from the context is 97.8~F1.~\footnote{To compute the upper bound, if abstractive answer is exactly ``yes'', ``no'', or ``unknown'', we consider the upper bound to be 100. Otherwise, we use the \coqa~evaluation script to find a span in the context that has maximum F1 with respect to the abstractive answer. }

\begin{table}[t]
\begin{center}
 \small

\begin{tabular}{l|c}
 & Overall F1
\\ \midrule
DrQA (Extractive) &  54.7 \\
DrQA + PGNet (Abstractive) &  66.2  \\
\midrule
BiDAF++ w/ 0-ctx  &   63.4 \\
BiDAF++ w/ 3-ctx  &  {\bf 69.2} \\
\end{tabular}
\vspace{-7pt}
\caption{Development set performance by training BiDAF++~\cite{ChoiQuAC2018} models (extractive) on \coqa~data with handling yes/no and no-answer questions as in \quact. Despite being extractive, these models significantly outperform reported baselines, DrQA and DrQA + PGNet ~\cite{Reddy18}.}
\vspace{-7pt}
\label{tab:extractivecoqa}
\end{center}
\end{table}

\begin{table}[t]
 \small

    \centering
    \begin{tabular}{l|c|c|c}
    
     & in-F1 & out-F1 & F1  \\
    \midrule
    DrQA & 54.5 & 47.9 & 52.6\\
    DrQA + PGNet & 67.0 & 60.4 & 65.1 \\
    BiDAF++ w/ 3-ctx & {\bf 69.4} & {\bf 63.8} & {\bf 67.8} \\
    \end{tabular}
        \vspace{-7pt}
    \caption{Test set results on \coqa. We report in domain F1 (in-F1), out of domain F1 on two held out domains, Reddit and Science (out-F1) and the overall F1 (F1).  }
            \vspace{-7pt}

    \label{tab:coqa_test}
\end{table}
\section{New Extractive Baseline for \coqa}
\label{sec:baseline}
Our analysis strongly implies that beyond yes/no questions, abstractive behavior is not a significant component in either \quact~ or \coqa. As such, \quact~models can be trivially adapted to \coqa. 

We train a set of BiDAF++ baselines from the original \quact~ dataset release~\cite{ChoiQuAC2018} by optimizing the model to predict the span with maximum F1 overlap with respect to annotated abstractive answers.\footnote{We use the implementation on \url{http://allennlp.org}, and do not modify any hyper-parameters except the the maximum dialog length and that models were allowed to train up to 65 epochs.} 
If the abstractive answer is exactly ``yes'' or ``no'', we train the model to output the whole rationale span, and classify the question as yes/no with the appropriate answer. 
At evaluation time, if the model predicts a question is a yes/no question, instead of returning the extracted span, we simply return ``yes'' or ``no''.  

\paragraph{Results} Table~\ref{tab:extractivecoqa} and Table~\ref{tab:coqa_test} summarize our results for training BiDAF++ with varying contexts on \coqa.
Beyond the difference of underlying base question-answer models (DrQA~\cite{chen2017reading} vs. BiDAF~\cite{seo2016bidirectional} with self attention~\cite{clark2017simple}), BiDAF++ has two core differences with respect to DRQA+PGNet: (1) instead of appending previous questions and answers to input question tokens, BiDAF++ marks answers of previous questions directly on the context, and (2) BiDAF++ uses contextualized word embeddings through ELMo~\cite{peters2018deep}. These differences, in combination with appropriate handling of yes/no and unanswerable questions significantly improves on the existing extractive baseline {\bf (+14.2 F1)} and even on the existing abstractive baseline {\bf (+2.7 F1)}.


\begin{table}[t]
 \small

    \centering
    \begin{tabular}{l|c|c|c}
    
     & F1 & HEQQ & HEQD \\
    \midrule
    BiDAF++ w/ 2-ctx & 60.6    & 55.7  &  4.0  \\
    \midrule 
    Train \squad & 34.3  & 18.0 & 0.3 \\
    Train \coqa & 31.2  & 19.2 & 0.0  \\
    Ft from \squad & 62.6  & 58.3  & \bf{5.9}  \\
    Ft from \coqa & \bf{63.3}  & \bf{59.2} & 5.1 \\
    \end{tabular}
    \vspace{-7pt}
    \caption{Cross dataset transfer to \quact~development set. Models do not transfer directly (rows 3 and 4), but after fine tuning improve performance (rows 5 and 6).}
        \vspace{-7pt}

    \label{tab:quac_experiments}
\end{table}

\section{Cross-Dataset Experiments}
\label{sec:results}
In this section we consider whether models can benefit from transfer between \squad, \quact, and \coqa, and show that the datasets, while ineffective for direct transfer, can be used as pretraining.
In all experiments, we use BiDAF++, either with two context or no context, depending on if we are training for dialog settings or not, with default configurations.
Models are trained by initializing from other models trained on different datasets and we do not decrease initial learning rates from just training directly on the target dataset.
When \squad~is used to initialize models that use context, we randomly order questions in \squad~ and train as if questions were asked in the form of a dialog.~\footnote{Likely a better strategy exists but we would like to demonstrate transfer in the simplest way. We only report development numbers as these experiments are meant to be exploratory.}



\begin{table}[t]
 \small

    \centering
    \begin{tabular}{l|c}
    
     & In Domain F1   \\
    \midrule
    DrQA + PGNet & 66.2 \\
    BiDAF++ w/ 2-ctx & 67.6 \\
    \midrule 
    \squad & 41.4  \\
    \quact & 29.1  \\
    Ft from \squad & \bf{69.2}  \\
    Ft from \quact & 68.0 \\
    \end{tabular}
    \vspace{-7pt}
    \caption{Cross dataset transfer to \coqa~development set. Models do not transfer directly (rows 3 and 4), but after fine tuning improve performance (rows 5 and 6). For an explanation of why BiDAF++ outperforms DrQA + PGNet, see Section~\ref{sec:baseline}.}
        \vspace{-7pt}

    \label{tab:coqa_experiments}
\end{table}


\begin{table}[t]
 \small

    \centering
    \begin{tabular}{l|c|c}
    
     &F1 & EM   \\
    \midrule
    Baseline & 67.6 & 65.1\\
    BiDAF++ & 70.5 & 67.4\\
    \midrule 
    \coqa & 38.1  & 32.4 \\
    \quact & 25.4  & 16.8 \\
    Ft from \coqa & {\bf 72.5} & {\bf 69.4} \\
    Ft from \quact & 69.5 & 66.8\\
    \end{tabular}
    \vspace{-7pt}
    \caption{Cross dataset transfer to \squad~development set. BiDAF++~\cite{ChoiQuAC2018} outperforms the baseline, a different implementation of the same model~\cite{Rajpurkar2018KnowWY} likely because of better hyper parameter tuning.}
        \vspace{-7pt}

    \label{tab:squad2_experiments}
\end{table}

\vspace{-7pt}
\paragraph{Results} Tables 6-8 summarize our results. 
Across all of the datasets, BiDAF++ outperforms other baselines, and there exists at least one other dataset that significantly improves performance on a target dataset on average {\bf +2.1 F1}. 
Experiments do not support that direct transfer is possible.

\section{Related Work}\label{sec:related}
Other proposals exist other than the three we analyzed that expand on features in SQuAD~\cite{rajpurkar2016squad}. 
For example, maintaining question independence of context to reduce the role of string matching and having long context length~\cite{JoshiTriviaQA2017,Kocisk2017TheNR}, higher level reasoning ~\cite{MultiRC2018,clark2018think,yang2018hotpotqa}, multi-turn information seeking interactions, in either table settings~\cite{IyyerSQA2017,talmor2018web,saha2018complex}, regulation settings~\cite{Saeidi2018QuARC}, or Quiz Bowl settings~\cite{quizbowl}.  
Other work considers multi-modal contexts where interactions are a single turn ~\cite{tapaswi2016movieqa,antol2015vqa,lei2018tvqa} or multi-turn~\cite{das2017visual,pasunuru2018gamebased}.
These efforts contain alternative challenges than ones we analyze in this paper.

\section*{Acknowledgement}
We thank Eunsol Choi, Hsin-Yuan Huang, Mohit Iyyer, He He, Yejin Choi, Percy Liang, and Luke Zettlemoyer for their helpful discussions in formulating this work. Also, Siva Reddy and Danqi Chen for help evaluating on CoQA and all reviewers for their comments.

\bibliographystyle{bib/acl_natbib_nourl}
\bibliography{bib/journal-full,bib/main}

\end{document}